\documentclass{article}

    \usepackage[final]{neurips_2021}

\usepackage{epsfig}
\usepackage{graphicx}

\usepackage[utf8]{inputenc} 
\usepackage[T1]{fontenc}    
\usepackage{hyperref}       
\usepackage{url}            
\usepackage{booktabs}       
\usepackage{amsfonts}       
\usepackage{nicefrac}       
\usepackage{microtype}      
\usepackage{xcolor}         
\usepackage{times}

\usepackage{subcaption}

\usepackage{amsmath}
\usepackage{amssymb}

\title{Exploring Latent Dimensions of Crowd-sourced Creativity}

\author{\stepcounter{footnote}\vspace{1mm}Umut Kocasari$^{}$\thanks{Equal contribution. Author ordering determined by a coin flip.}
\hspace{0.75cm}
Alperen Bag$^{}$\footnotemark[2] 
\hspace{0.75cm}
Efehan Atici
\hspace{0.75cm}
Pinar Yanardag
\\
\hspace{1em} 
Bogazici University
\\
{\tt\small \{umut.kocasari,alperen.bag,efehan.atici\}@boun.edu.tr}\\
{\tt\small yanardag.pinar@gmail.com}
}

\begin{document}

\maketitle

\begin{abstract}
 Recently,  the discovery of interpretable directions in the latent spaces of pre-trained GANs has become a popular topic. While existing works mostly consider directions for semantic image manipulations, we focus on an abstract property: \textit{creativity}. Can we manipulate an image to be more or less \textit{creative}? We build our work on the largest AI-based creativity platform, Artbreeder,  where users can \textit{generate}  images using pre-trained GAN models. We explore the latent dimensions of images generated on this platform and present a novel framework for manipulating images to make them more \textit{creative}.  Our code and dataset are available at \url{http://github.com/catlab-team/latentcreative}.
\end{abstract}

\section{Introduction}

One of the most popular platforms that collaboratively take advantage of generative adversarial networks is Artbreeder \cite{Artbreeder}. Artbreeder helps users create new images using a BigGAN \cite{BigGAN} model, where users can adjust parameters or mix different images to generate new ones. Artbreeder platform is investigated by several recent work \cite{rafner2020power, hertzmann2020visual, wang2020toward, gokay2021graph2pix}. In this study, we explore the manipulation of the latent space for \textit{creativity} using a large dataset from the Artbreeder platform. We use this data as a proxy to build an assessor model that indicates the extent of creativity for a given image. Our approach modifies the image semantics by shifting the latent code towards creativity by a certain amount to make it more or less \textit{creative}.

\section{Methodology}
\label{sec:methodology}
In this paper, we use a pre-trained BigGAN model to manipulate images towards creativity and build our work on the GANalyze \cite{goetschalckx2019ganalyze} model. Given a generator $G$, a class vector $y$, a noise vector $z$, and an assessor function $A$, the GANalyze model solves the following problem: $ \mathcal{L}(\theta) = \mathbb{E}_{\mathbf{z},\mathbf{y},\alpha} [ A(G(T_{\theta}(\mathbf{z}, \alpha), \mathbf{y})) - (A(G(\mathbf{z}, \mathbf{y})) + \alpha))^2 ] $, where $\alpha$ is a scalar value representing the degree of manipulation, $\theta$ is the desired direction, and $T$ is the transformation function defined as $T_\theta (\mathbf{z},\alpha) = \mathbf{z} + \alpha\theta$ that moves the input $\mathbf{z}$ along the  direction $\theta$. In this work, we extend the GANalyze framework where we use a neural network that uses noise drawn from a distribution $\mathcal{N}(0, 1)$ which learns to map an input to different but functionally related (e.g., more creative) outputs:

\begin{equation}
 \mathcal{L}_z(\theta) = \mathbb{E}_{\mathbf{z},\mathbf{y},\alpha} [ (A(G(F_{z}(\mathbf{z}, \alpha), \mathbf{y})))  -  (A(G(\mathbf{z}, \mathbf{y})) + \alpha))^2 ] 
\end{equation}

where the first term represents the score of the modified image after applying the function $F$ with parameters $\mathbf{z}$, $\mathbf{y}$, and $\alpha$, and the second term simply represents the score of the original image increased or decreased by $\alpha$. $F_z$ computes a diverse direction $\theta$ with a noise $ \epsilon$ as $F_{z}(\mathbf{z}, \alpha) = \mathbf{z} + \alpha \cdot \mathbf{NN}(\mathbf{z}, \epsilon)$. Moreover, we also learn a direction for class vectors as follows:

\begin{equation}
 \mathcal{L}(\theta) = \mathbb{E}_{\mathbf{z},\mathbf{y},\alpha} [ A(G(F_{z}(\mathbf{z}, \alpha), F_{y}(\mathbf{y}, \alpha)))  -  (A(G(\mathbf{z}, \mathbf{y})) + \alpha))^2 ] 
\end{equation}

 where $F_y$ is calculated as $
 F_{y}(\mathbf{y}, \alpha) = \mathbf{y} + \alpha \cdot \mathbf{NN}(\mathbf{y}, \epsilon)$ where NN  is a two layer neural  network.

 \section{Results}
 
\begin{figure*}[t!]
\minipage{0.25\textwidth}
  \includegraphics[width=\linewidth]{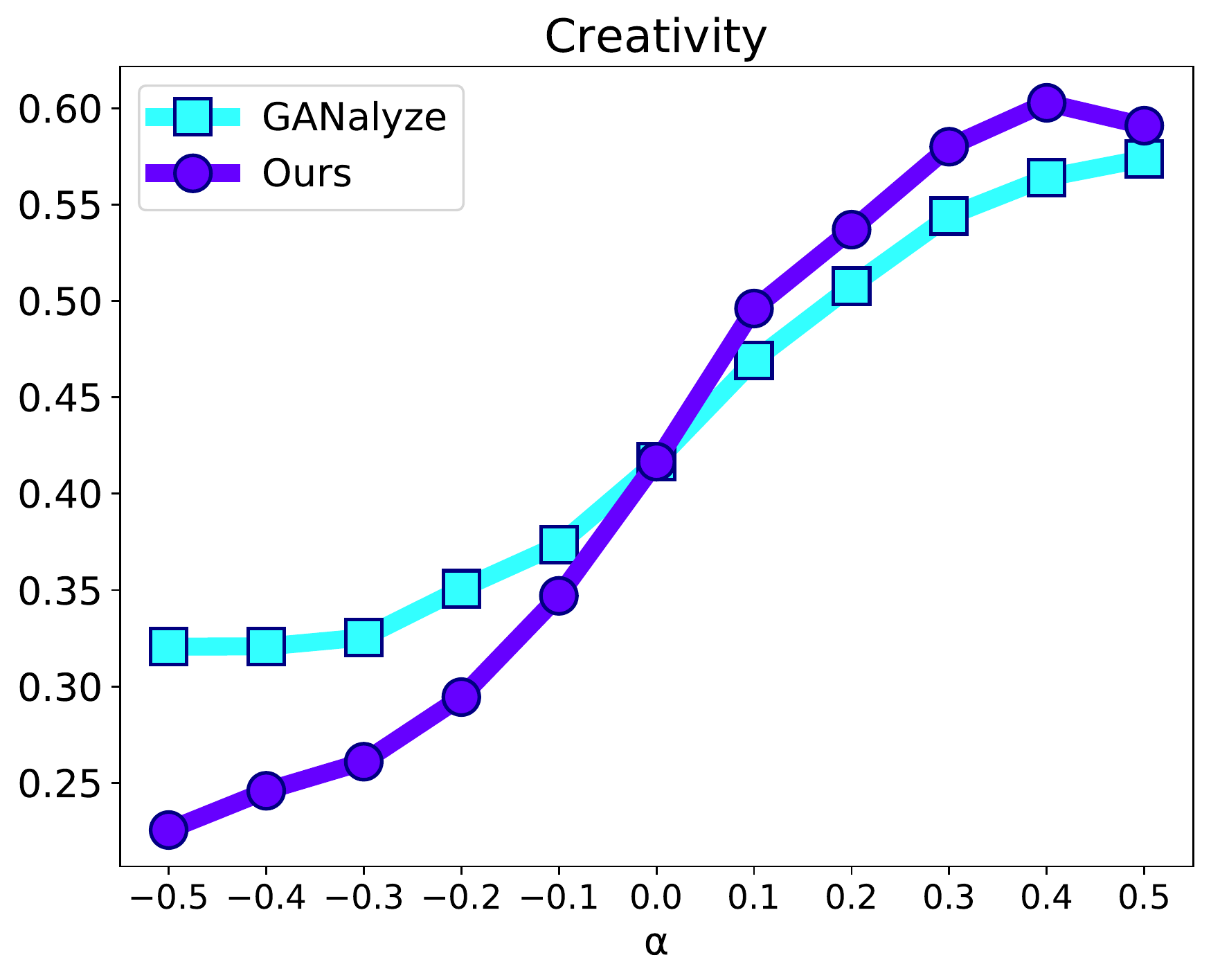}
  \caption*{(a)} 
\endminipage\hfill
\minipage{0.25\textwidth}
  \includegraphics[width=\linewidth]{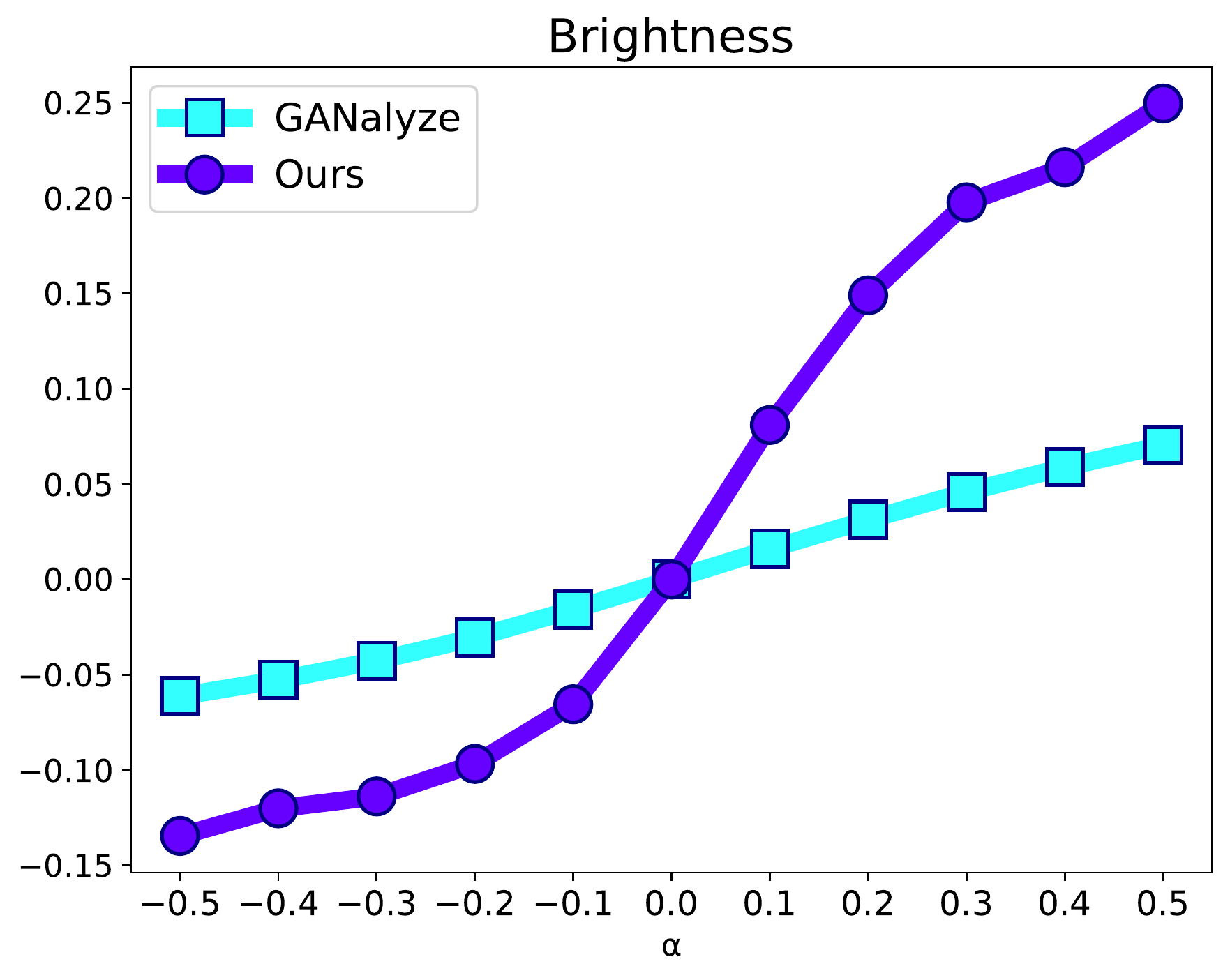}
  \caption*{(b)} 
\endminipage\hfill
\minipage{0.25\textwidth}
  \includegraphics[width=\linewidth]{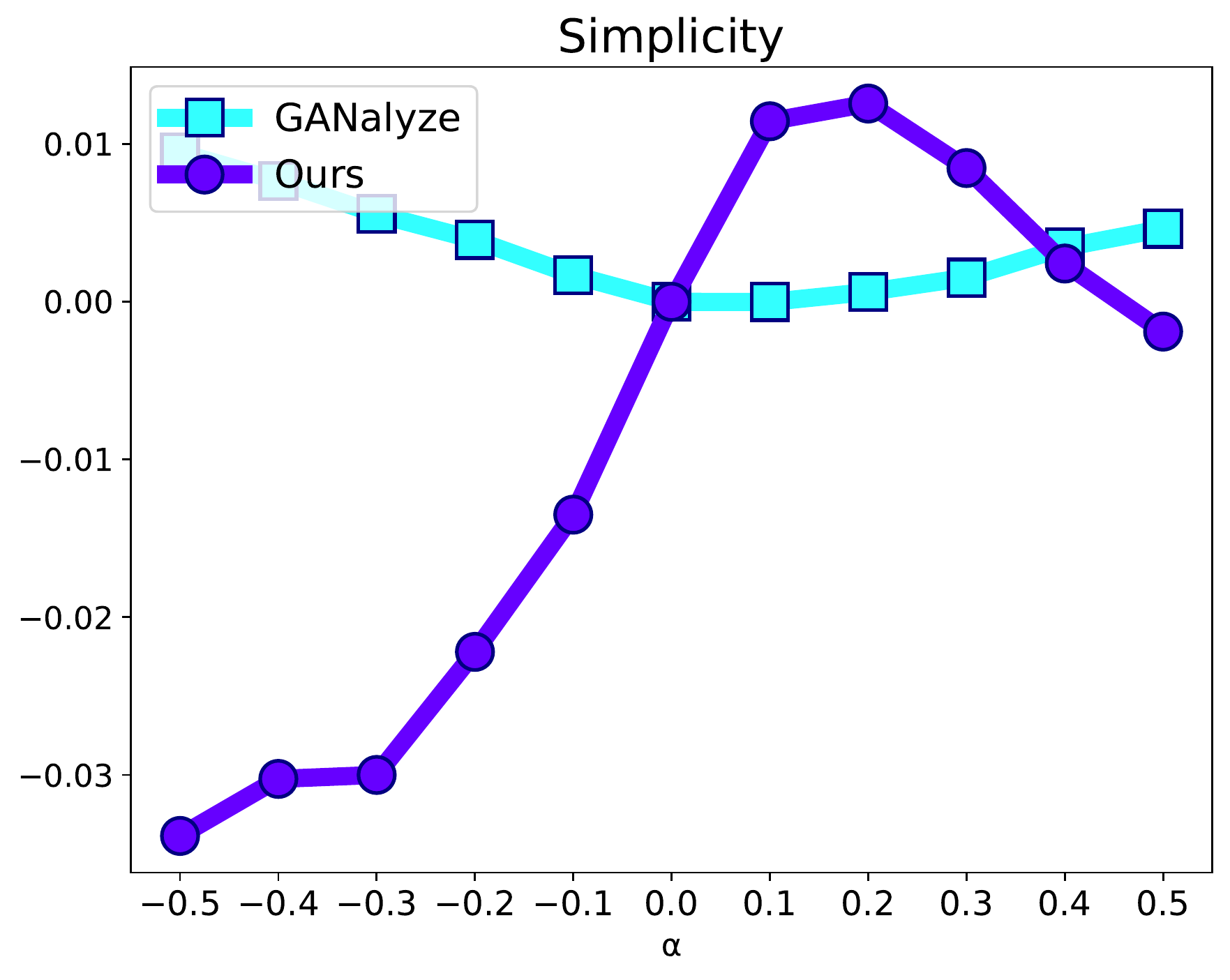}
  \caption*{(c)} 
\endminipage\hfill
\minipage{0.25\textwidth}%
  \includegraphics[width=\linewidth]{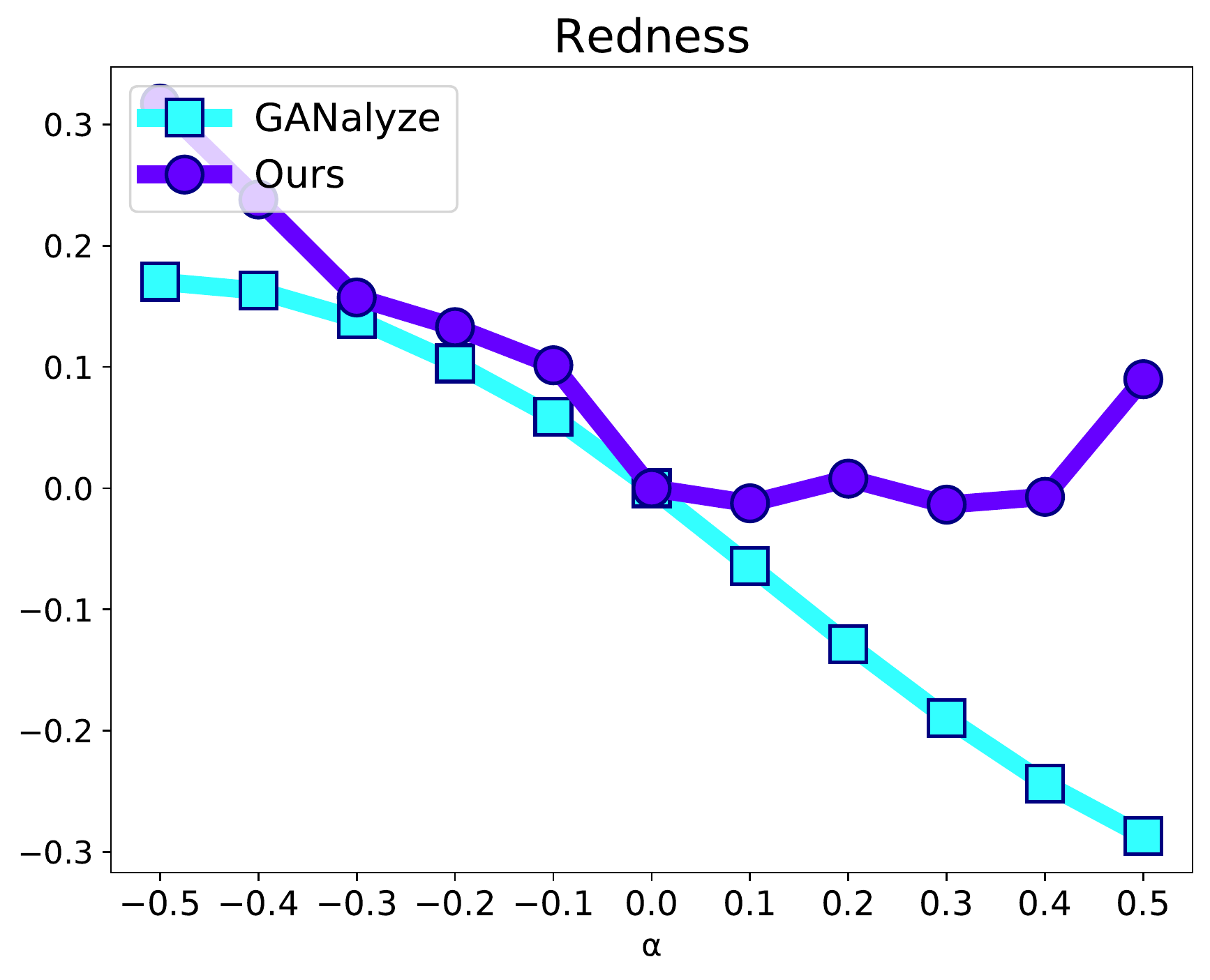}
  \caption*{(d)} 
\endminipage

\minipage{0.25\textwidth}
  \includegraphics[width=\linewidth]{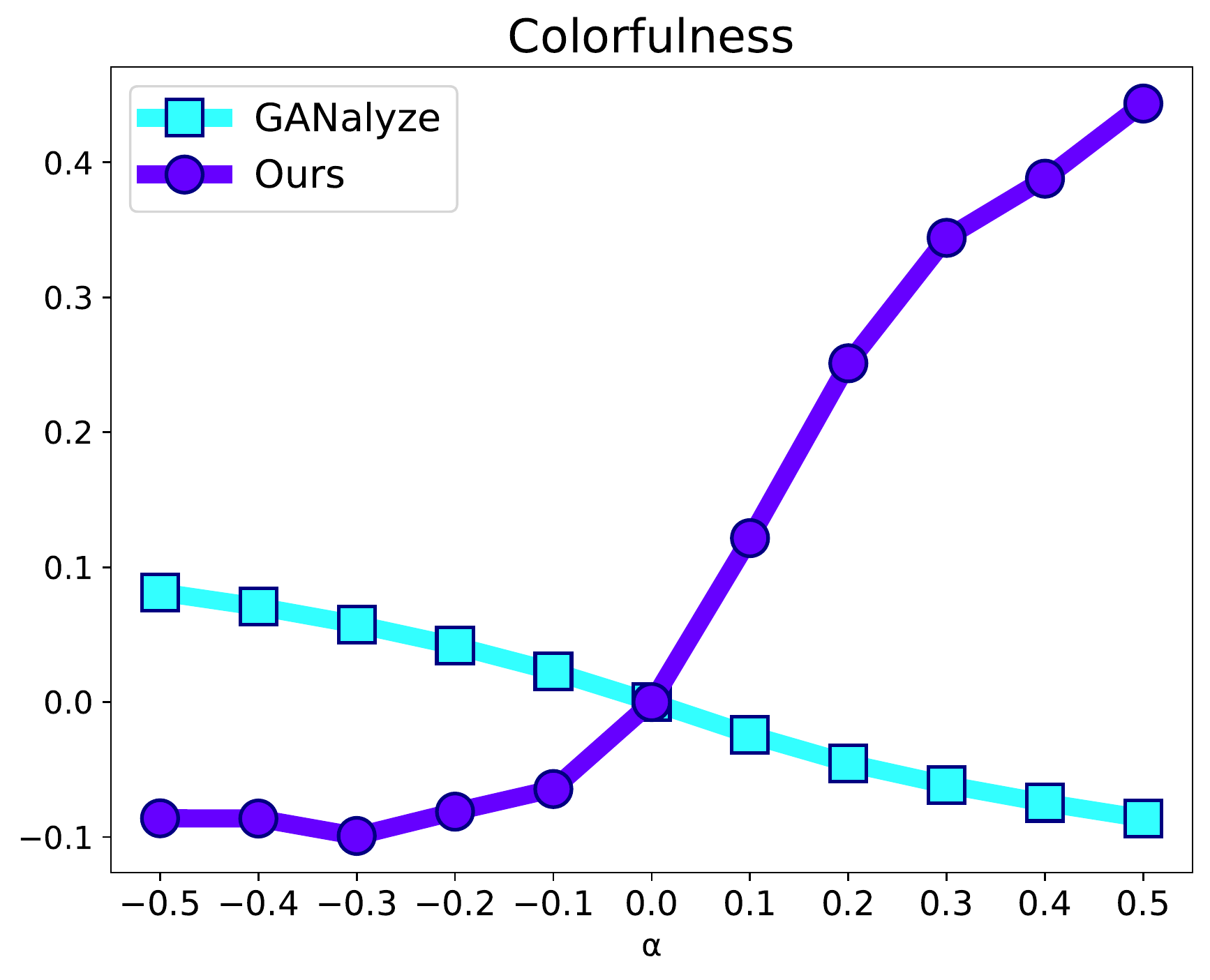}
  \caption*{(e)} 
\endminipage\hfill
\minipage{0.25\textwidth}
  \includegraphics[width=\linewidth]{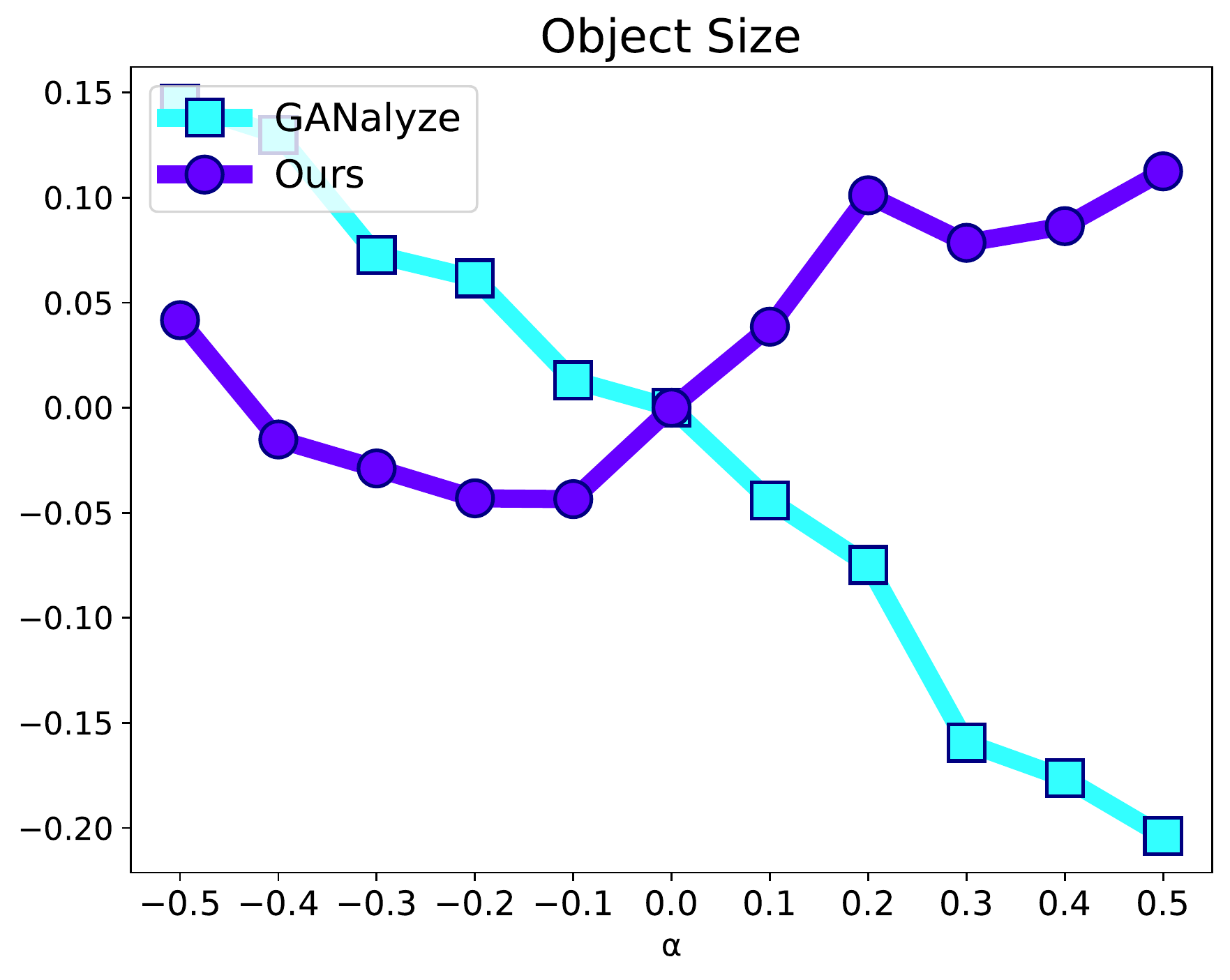}
  \caption*{(f)} 
\endminipage\hfill
\minipage{0.25\textwidth}
  \includegraphics[width=\linewidth]{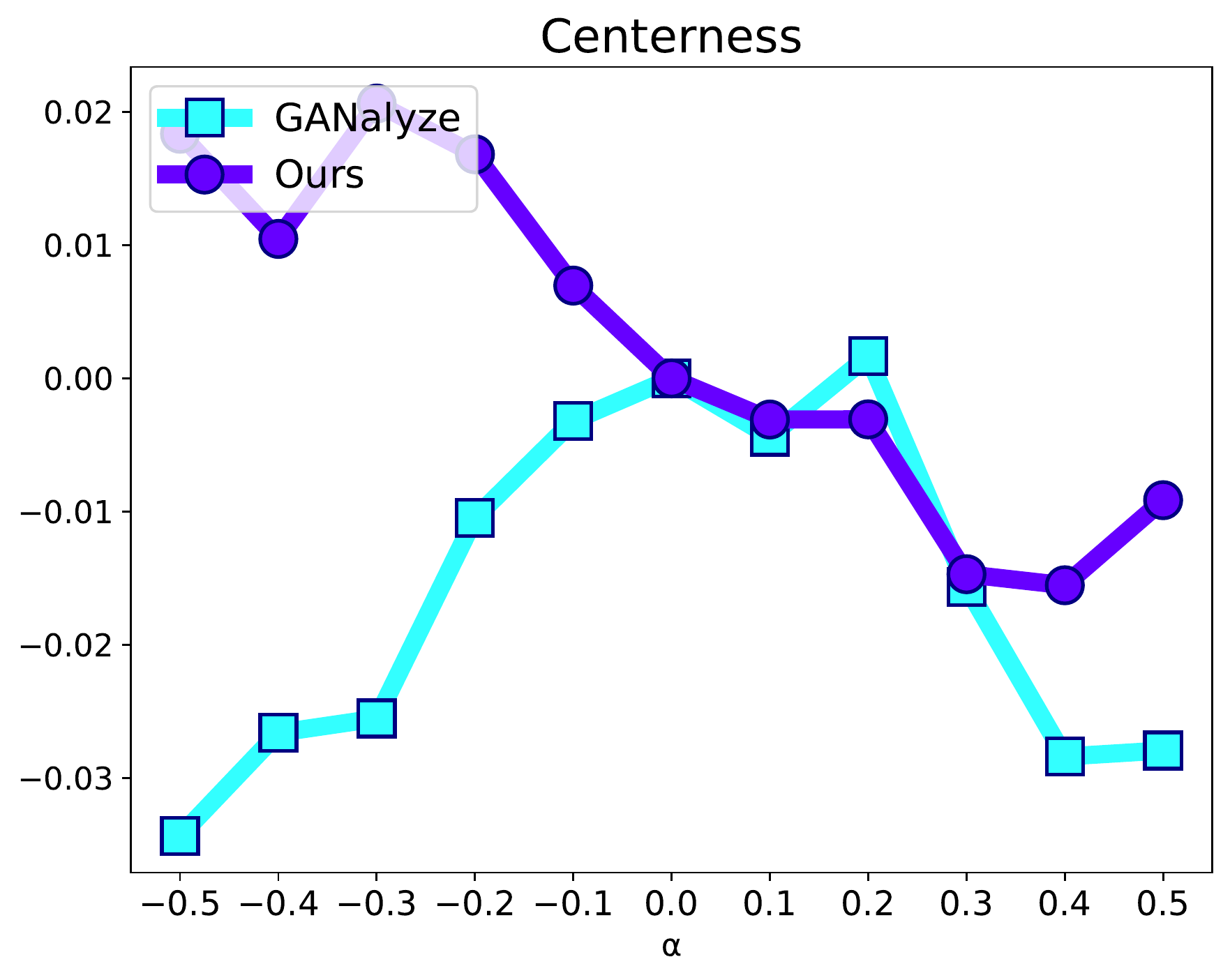}
  \caption*{(g)} 
\endminipage\hfill
\minipage{0.25\textwidth}%
  \includegraphics[width=\linewidth]{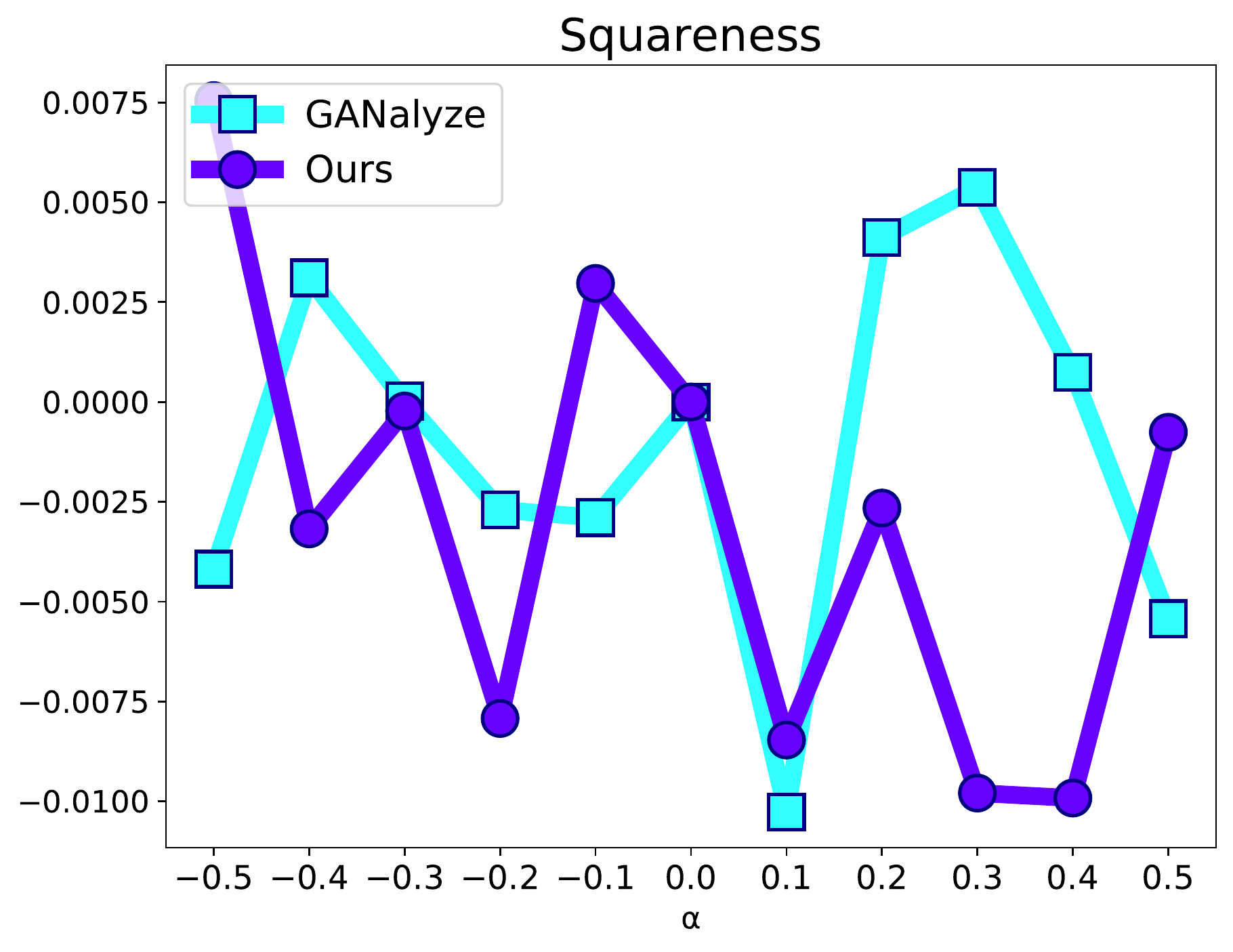}
  \caption*{(h)} 
\endminipage

  \caption{Our method learns a better manipulation of the latent space compared to GANalyze \cite{goetschalckx2019ganalyze}. For each figure, the $x$-axis represents the $\alpha$-value and the $y$-axis represents the mean.}
  \label{fig:exp_cond}
\end{figure*}

\begin{figure}
  \centering
    \includegraphics[width=1\linewidth]{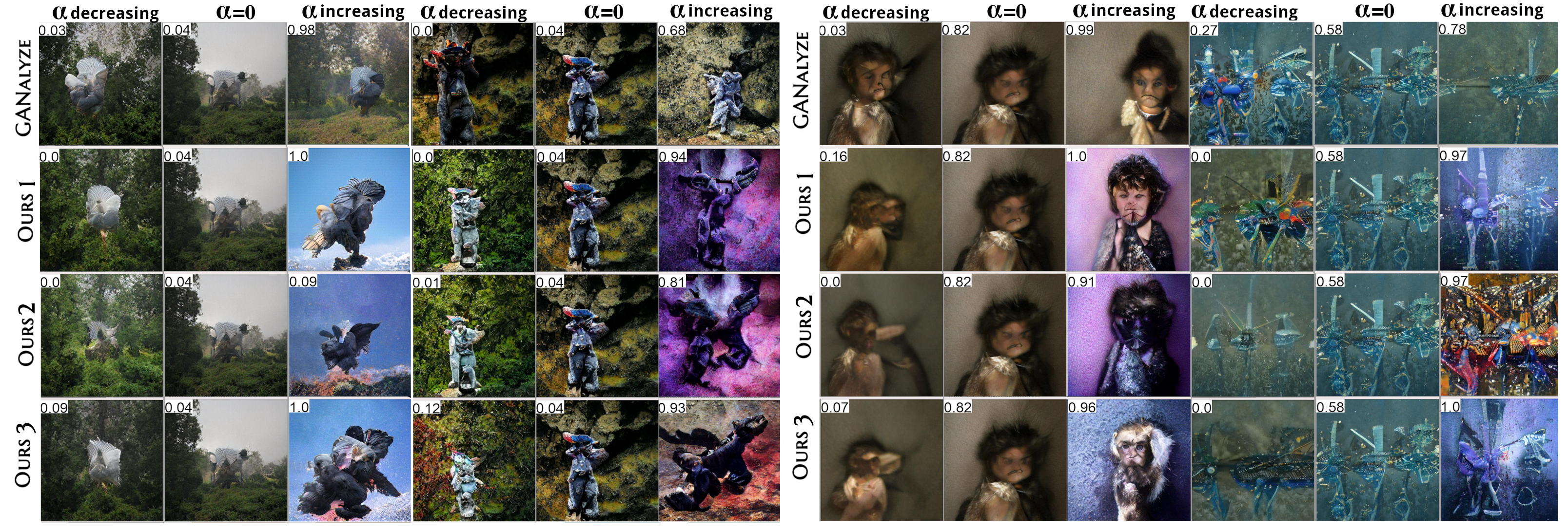}
  \caption{Generated images created by our model and GANalyze \cite{goetschalckx2019ganalyze}.  }
  \label{fig:sample_results}
\end{figure}

We first investigate whether our model learns to navigate the latent space in such a way that it can increase or decrease creativity by a given $\alpha$ value. As can be seen from Figure \ref{fig:exp_cond} (a), our model achieves a much lower creativity score for low $\alpha$ values such as $\alpha=-0.5$, while simultaneously achieving higher creativity for $\alpha=0.4$ and $\alpha=0.5$ compared to \cite{goetschalckx2019ganalyze}. Next, we investigate what kind of image factors are changed to achieve these improvements. Following \cite{goetschalckx2019ganalyze}, we examined \textit{redness, colorfulness, brightness, simplicity, squareness, centeredness} and \textit{object size} (see Appendix for detailed definitions of each property). We observe that both methods follow similar trends except \textit{colorfulness} and \textit{object size}. In particular, our method prefers to increase colorfulness and object size with increasing $\alpha$ values. Figure \ref{fig:sample_results} shows some of the samples generated by our method and baseline \cite{goetschalckx2019ganalyze}. As can be seen from the visual results, our method is capable of performing several diverse manipulations on the input images.

\section{Limitations and Broader Impact}
\label{sec:limitations}
Our method is limited to manipulating GAN-generated images. As with any image synthesis tool, our method faces similar concerns and dangers of misuse if it can be applied to images of people or faces for malicious purposes, as described in \cite{korshunov2018deepfakes}. Our work also sheds light on the understanding of creativity and opens up possibilities to improve human-made art through GAN applications. On the other hand, we point out that using GAN-generated images as a proxy for creativity may bring its own limitations.

{\small
\bibliographystyle{ieee_fullname}
\bibliography{neurips_2021}

\begin{thebibliography}{10}\itemsep=-1pt

\bibitem{Artbreeder}
Artbreeder.com.

\bibitem{BigGAN}
Andrew Brock, Jeff Donahue, and Karen Simonyan.
\newblock Large scale {GAN} training for high fidelity natural image synthesis.
\newblock {\em CoRR}, abs/1809.11096, 2018.

\bibitem{goetschalckx2019ganalyze}
Lore Goetschalckx, Alex Andonian, Aude Oliva, and Phillip Isola.
\newblock Ganalyze: Toward visual definitions of cognitive image properties.
\newblock In {\em Proceedings of the IEEE/CVF International Conference on
  Computer Vision}, pages 5744--5753, 2019.

\bibitem{gokay2021graph2pix}
Dilara Gokay, Enis Simsar, Efehan Atici, Alper Ahmetoglu, Atif~Emre Yuksel, and
  Pinar Yanardag.
\newblock Graph2pix: A graph-based image to image translation framework.
\newblock In {\em Proceedings of the IEEE/CVF International Conference on
  Computer Vision}, pages 2001--2010, 2021.

\bibitem{hasler2003measuring}
David Hasler and Sabine~E Suesstrunk.
\newblock Measuring colorfulness in natural images.
\newblock In {\em Human vision and electronic imaging VIII}, volume 5007, pages
  87--95. International Society for Optics and Photonics, 2003.

\bibitem{he2017mask}
Kaiming He, Georgia Gkioxari, Piotr Doll{\'a}r, and Ross Girshick.
\newblock Mask r-cnn.
\newblock In {\em Proceedings of the IEEE international conference on computer
  vision}, pages 2961--2969, 2017.

\bibitem{hertzmann2020visual}
Aaron Hertzmann.
\newblock Visual indeterminacy in gan art.
\newblock {\em Leonardo}, 53(4):424--428, 2020.

\bibitem{korshunov2018deepfakes}
Pavel Korshunov and S{\'e}bastien Marcel.
\newblock Deepfakes: a new threat to face recognition? assessment and
  detection.
\newblock {\em arXiv preprint arXiv:1812.08685}, 2018.

\bibitem{PCAAnalysis}
Andrzej Maćkiewicz and Waldemar Ratajczak.
\newblock Principal components analysis (pca).
\newblock {\em Computers \& Geosciences}, 19(3):303 -- 342, 1993.

\bibitem{rafner2020power}
Janet Rafner, Lotte Philipsen, Sebastian Risi, Joel Simon, and Jacob Sherson.
\newblock The power of pictures: using ml assisted image generation to engage
  the crowd in complex socioscientific problems.
\newblock {\em arXiv preprint arXiv:2010.12324}, 2020.

\bibitem{TSNE}
Laurens van~der Maaten and Geoffrey Hinton.
\newblock Viualizing data using t-sne.
\newblock {\em Journal of Machine Learning Research}, 9:2579--2605, 11 2008.

\bibitem{wang2020toward}
Xi Wang, Zoya Bylinskii, Aaron Hertzmann, and Robert Pepperell.
\newblock Toward quantifying ambiguities in artistic images.
\newblock {\em ACM Transactions on Applied Perception (TAP)}, 17(4):1--10,
  2020.

\end{thebibliography}
}

\section{Appendix}

 Following \cite{goetschalckx2019ganalyze}, we explored \textit{Redness, Colorfulness, Brightness, Simplicity, Squareness, Centeredness} and \textit{Object Size}. 
 
\begin{itemize} 
\item \textbf{Redness} is calculated as the normalized number of red pixels.
\item \textbf{Colorfulness} is calculated by the metric from \cite{hasler2003measuring}. 
\item \textbf{Brightness} is calculated as the average pixel value after the grayscale transformation.
\item \textbf{Simplicity} is computed as a histogram of pixel intensity used to measure entropy. \item \textbf{Object Size} Following \cite{goetschalckx2019ganalyze}, we used Mask R-CNN \cite{he2017mask} to create a segmentation mask at the instance level of the generated image. For \textit{Object Size}, we used the difference in the mask area when the $\alpha$ value increases or decreases. 
\item  \textbf{Centeredness} is calculated as the deviation between the centroid of the mask from the center of the frame. 
\item \textbf{Squareness} is calculated as the ratio of the length of the minor and major axes of an ellipse with the same normalized central moments as the mask. 
\end{itemize}
 
Figure \ref{fig:exp_cond} shows the comparison between \cite{goetschalckx2019ganalyze} and our method. For Brightness \ref{fig:exp_cond} (b), we can see that both methods increase the brightness with increasing value of $\alpha$, while our method prefers to increase it with a larger distance. As can be seen from Figure \ref{fig:exp_cond} (c), our method increases the simplicity up to a certain $\alpha$ value and decreases it at large $\alpha$ values such as $\alpha=0.4$ and $\alpha=0.5$. We believe that this is due to the fact that images with higher creativity contain more complex objects. On the other hand, we see that the other method decreases the simplicity until $\alpha=0$ and then increases it slightly as $\alpha$ increases. For Redness (Figure \ref{fig:exp_cond} (d)), both methods prefer to decrease it when the $\alpha$ value becomes higher. For Colorfulness (Figure \ref{fig:exp_cond} (e)) GANalyze consistently decreases the value, while our method prefers to increase it at a much higher rate. For Object Size (Figure \ref{fig:exp_cond} (f)), we can see that GANalyze consistently decreases the object size when the $\alpha$ value becomes larger, while our method follows a nonlinear behavior. A similar behavior is seen for Centeredness (Figure \ref{fig:exp_cond} (g)), where GANalyze increases it until $\alpha=0$ and then decreases it. Our method prefers to decrease the centeredness as $\alpha$ increases. For Squaredness (Figure \ref{fig:exp_cond} (h)), both methods follow a non-linear trend.
 
\subsection{Experimental Setup}
\label{sec:experimental_setup}
All methods are trained for $400K$ iterations, and we use a batch size of 8,  Rectified Adam optimizer, and a learning rate of $1e-3$.  Following \cite{goetschalckx2019ganalyze}, we randomly select $1000$ images from the Artbreeder website. Each $z$ vector is paired with 11 different $\alpha$ values $[-0.5,-0.4,-0.3,-0.2,-0.1,0,0.1,0.2,0.3,0.4,0.5]$, where $\alpha = 0$ recovers the original image $G(z,y)$. Thus, our final test set consists of $1500 \times 11$ images, for a total of $16,500$ images.

\subsection{Creativity Assessor}
To manipulate the images towards creativity, we need a scoring model $A$ that takes an image and returns a score indicating how \textit{creative} the image is. However, scoring creativity is a challenging problem and there is no easy way to directly infer the creativity of an image. Therefore, we use the images collected by Artbreeder as a proxy for creativity, as users try to generate more creative images by combining existing images ones. Thus, we hypothesize that images with a high number of ancestors are more creative than images with a low number of ancestors. To test this hypothesis, we conduct an evaluation study in which we randomly show participants one of two images and ask them which image is more creative: an image generated directly by the BigGAN model and an image with more than 100 ancestors. We randomly selected $15$ images and asked $15$ participants to vote on which image was more creative. Our perceptual study confirms that participants indeed find images with more ancestors more creative than images without ancestors: they find the images with more ancestors $89.7\%$ more creative than the images without ancestors. Although creativity is a very complex property, we believe that our approach represents a simple step towards understanding collective creativity. 

Our next step was to teach this notion to a classifier. To do this, we collected 48,000 images with more than 100 ancestors, 24,000 images with 0 ancestors, and generated 24,000 images from BigGAN by giving randomly sampled $z$-vectors and one-hot $y$-vectors. We labeled the first half as creative due to the high number of ancestors, meaning that they attract people more than other images, and labeled the other half as non-creative because it is just an image of an object belonging to an Imagenet class. We used the EfficientNet-B0 architecture to train our model and obtained an accuracy of 92\% on our validation set (80/20 random split). This shows that there is a significant difference between the two classes of images captured by our CNN model (see Section \ref{sec:experimental_setup} for more details).

\subsection {Data Collection}
We collected images from Artbreeder at regular intervals by crawling the 'Trending' section of the 'General' category daily over a four-week period \footnote{Note that the images on Artbreeder are in the public domain (CC0) and we obtained permission from the owner to crawl the images.}. The resolution of the images provided by Artbreeder varies from $512 \times 512$ to $1024 \times 1024$. To ensure that we have a consistent dataset, we resized all images to $512 \times 512$.  

\subsection{Data Exploration}
\label{sec:exploration}

\begin{figure*}[t] \centering \includegraphics[width=1\columnwidth]{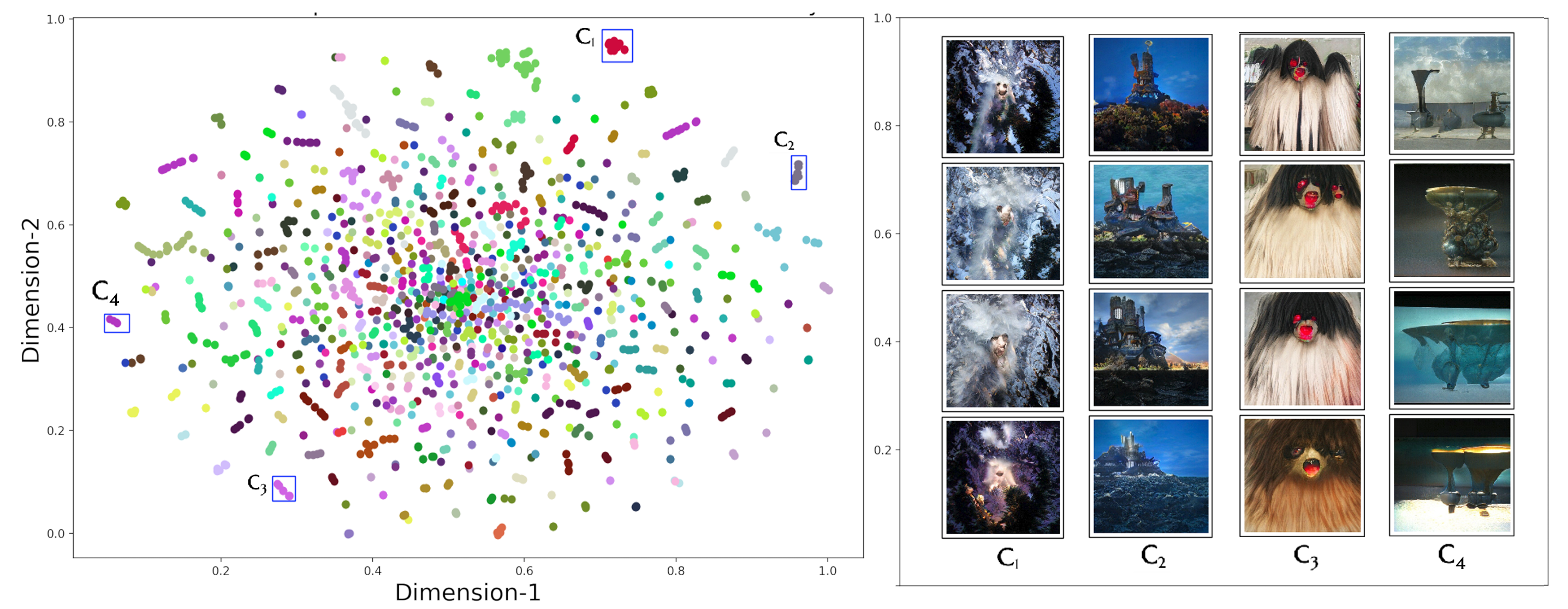} \caption{Examining generated images based on user groups. The images are mapped in a 2-dimensional space, preserving their proximity by t-SNE \cite{TSNE}, where the color of the points indicates the user who created the image. The clusters labeled $C_{1}$ and the points \dots $C_{4}$ on the left show selected groups of images, whose corresponding images can be seen on the right.} \label{fig:tsne}
\end{figure*}

To gain insight into the creative crowd behind Artbreeder, we visually explore the latent space of images and users. We obtained the latent vectors provided by Artbreeder and project each vector as a point in 2-dimensional space. To achieve this, we first reduced the dimensionality of the latent vectors using principal component analysis (PCA) \cite{PCAAnalysis}. After reducing the dimensions of the latent vectors, we apply t-Distributed Stochastic Neighbor Embedding \cite{TSNE} for two-dimensional representation, preserving the closeness in the high-dimensional space. Figure \ref{fig:tsne} shows a set of selected image groups with their corresponding visual content, where the colors represent different users. For clarity, only a subset of users who created between 50 and 100 images are shown. While the middle of the figure shows that similar images can be created by different users, it can also be observed that certain users have a particular style that distinguishes them from others (like $C_{1}$). 

\section{Acknowledgements}
This publication has been produced benefiting from the 2232 International Fellowship for Outstanding Researchers Program of TUBITAK (Project No:118c321). We also acknowledge the support of NVIDIA Corporation through the donation of the TITAN X GPU and GCP research credits from Google. We thank to Joel Simon for his help on collecting data from Artbreeder platform.

\end{document}